\documentclass{article}
\usepackage{spconf,amsmath,graphicx}
\usepackage{subfigure}

\usepackage{algorithm}
\usepackage{algorithmic}
\usepackage{tabularx}

\title{Attr-Int: A Simple and Effective Entity Alignment Framework for Heterogeneous Knowledge Graphs}
\name{Linyan Yang\qquad Jingwei Cheng$^{\star}$\thanks{Corresponding Author. Email: chengjingwei@mail.neu.edu.cn. The work is supported by the National Natural Science Foundation of China (62276057), and Sponsored by CAAI-MindSpore Open Fund, developed on OpenI Community.}\qquad Chuanhao Xu\qquad Xihao Wang\qquad Jiayi Li\qquad Fu Zhang}
\address{School of Computer Science and Engineering, Northeastern University}
\begin{document}
%
\maketitle
\begin{abstract}
Entity alignment (EA) refers to the task of linking entities in different knowledge graphs (KGs). Existing EA methods rely heavily on structural isomorphism. However, in real-world KGs, aligned entities usually have non-isomorphic neighborhood structures, which paralyses the application of these structure-dependent methods. In this paper, we investigate and tackle the problem of entity alignment between heterogeneous KGs. First, we propose two new benchmarks to closely simulate real-world EA scenarios of heterogeneity. Then we conduct extensive experiments to evaluate the performance of representative EA methods on the new benchmarks. Finally, we propose a simple and effective entity alignment framework called Attr-Int, in which innovative attribute information interaction methods can be seamlessly integrated with any embedding encoder for entity alignment, improving the performance of existing entity alignment techniques. Experiments demonstrate that our framework outperforms the state-of-the-art approaches on two new benchmarks.

\end{abstract}
\section{Introduction}
\label{sec:intro}

Recently, knowledge graphs (KGs) have been built and applied in various domains. However, most KGs are constructed by different organizations and individuals, which inevitably brings about heterogeneity problems. Knowledge fusion \cite{shen2022comprehensive,pp} aims to align and merge heterogeneous and redundant information in KGs to form globally unified knowledge identification and association. Entity Alignment (EA) has been the key technique in knowledge graph fusion. The main purpose of EA is to find equivalent entities among different KGs.

The methods based on knowledge representation learning have gained their popularity in solving the entity alignment problem.  
These methods assume that equivalent entities from various KGs posses similar neighborhood structures (i.e., isomorphism). However, it is not always the case as different KGs may be highly heterogeneous \cite{tang2020bert}. 
Therefore, attempts have been made to alleviate heterogeneity by using the attention mechanism to assign different weights to the relations between entities \cite{mao2020mraea,wu2019relation} or by ignoring different neighbors that have a negative impact on the alignment task \cite{wu2020neighborhood,cao2019multi,li2019semi}.
The underlying mechanism of these methods to deal with heterogeneity is to transform heterogeneous problems into isomorphic problems, and lack of consideration for the cases that cannot be transformed into isomorphic problems. 
Furthermore, recent studies have observed that attribute triples can also provide crucial alignment signals \cite{liu2020exploring}. 
Most of these methods leverage attribute information by learning attribute name, type and value embeddings, and combine them with structural embeddings to obtain the alignment results \cite{sun2017cross,wang2018cross,chen2020jarka,zhang2019multi,trisedya2019entity,wang2020knowledge,tang2020bert,zhong2022semantics,zhang2023improving}. Due to the complexity and diversity of attribute information, there is a great memory and computing power overhead.

To deal with these challenges, we investigate and tackle the problem of entity alignment between heterogeneous KGs. First, we propose two new benchmarks to closely simulate real-world EA scenarios of heterogeneity. Then, we propose a simple and effective entity alignment framework (Attr-Int), in which innovative attribute information interaction methods can be seamlessly integrated with any embedding encoder in entity alignment, improving the performance of original entity alignment techniques. It is worth noting that in our framework, the utilization of attribute information does not require embedding learning, nor does it require any prior alignment of attributes.

In summary, the main contributions are as follows:
\begin{itemize}
    \item We propose two new benchmarks in which entity pairs share less identical neighbors to reproduce the scenario of EA between heterogeneous KGs, based on DBP15K \cite{sun2017cross} and OpenEA \cite{sun2020benchmarking} benchmarks.
    \item We propose an entity alignment framework called Attr-Int that integrates attribute information in a simple and effective way
    . In this framework, we provide attribute utilization method, \emph{uniqueness of attribute values}, and two interaction methods, \emph{result correction based on attribute information discrimination} and \emph{matrix combination based on parameter search}.
    \item We conduct extensive experiments to evaluate representative EA methods on the new benchmarks. Comprehensive experiments demonstrate our framework outperforms the state-of-the-art approaches. 
\end{itemize}

\section{Problem Definition}

\textbf{Definition 1} (Knowledge Graph). A knowledge graph (KG) is denoted as $G = (E, R, A, V, T_r, T_a)$, where $E=\{e_1, e_2,\ldots , e_m\}$, $R=\{r_1, r_2,\ldots , r_n\}$, $A=\{a_1, a_2,\ldots , a_p\}$, $V=\{v_1, v_2,\ldots ,, v_q\}$, $T_r \subseteq E\times R\times E$ and $T_a \subseteq E\times A\times V$ represent entity set, relation set, attribute set, attribute value set, relation triple set and attribute triple set respectively, and $m$, $n$, $p$, $q$ are the numbers of entities, relations, attributes, and attribute values respectively.

\textbf{Definition 2} (Entity Alignment in KGs). Given a source KG 
$G^1 = \{E^1, R^1, A^1, V^1, T_r^1, T_a^1\}$, and a target KG $G^2 = \{E^2, R^2, A^2, V^2, T_r^2, T_a^2\}$, the aligned entity pairs (training set) is denoted as $P = \{(e_i^1, e_j^2) | e_i^1 \in E^1, e_j^2 \in E^2, e_i^1\equiv e_j^2\}$, where $\equiv $ stands for equivalence, i.e., the entity $e_i^1$ and the entity $e_j^2$ refer to the same thing in the real world. 
The goal of the entity alignment task is to find remaining equivalent entity pairs of these two KGs. 

\textbf{Definition 3} (Coverage Rate of Entity Pairs). Let $(e_i^1, e_j^2)$ be an entity pair,  $N(e_i^1)$ and $N(e_j^2)$ be the sets of neighboring entities of $e_i^1$ and $e_j^2$ respectively, then the coverage rate $C_{(e_i^1, e_j^2)}$ of the entity pair $(e_i^1, e_j^2)$ is calculated by $ C_{(e_i^1, e_j^2)}=|N(e_i^1)\cap N(e_j^2)|/\min\big(|N(e_i^1)|,|N(e_j^2)|\big)$, where $|\cdot|$ represents the size of a set.

\section{Methodology}
Our proposed Attr-Int framework is shown in Figure 1, which consists of three major components: \textit{encoder}, \textit{attribute similarity module} and \textit{interaction module}. First, entity embeddings are obtained from the encoder, from which the similarities between entities are calculated, and the preliminary entity alignment result is obtained.
Then, the attribute similarity matrix is calculated by the attribute similarity module, and the entity alignment result based on attribute information is obtained.
Finally, the interaction module carries out sufficient cross-learning \cite{yuan2018iterative} on the above two results to obtain the final result. It is worth noting that, in the attribute similarity module, unlike previous methods leveraging attribute information, there is no need to learn attribute embedding or pre-align any attributes. 

\begin{figure}[h]
	\centering
	\includegraphics[width=8cm]{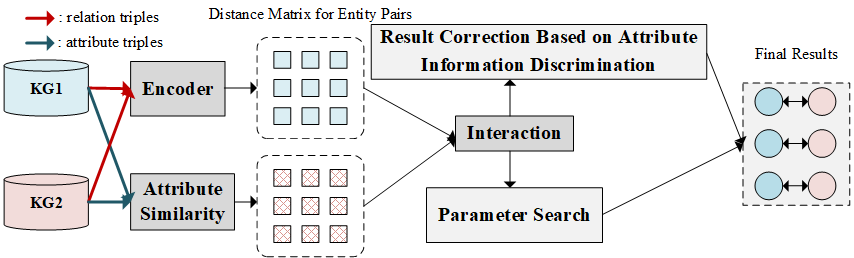}
	\caption{Overall architecture of proposed framework.} 
	\label{fig:4}
\end{figure}

\subsection{Encoder}
Our framework can incorporate any existing entity alignment technique. For the sake of generality, we consider the representative technique (RDGCN \cite{wu2019relation}) as the encoder in our framework. In order to better incorporate relation information into entity embeddings, RDGCN \cite{wu2019relation}  first constructs a dual relation graph for the input KG, in which vertices denote the relations in the original graph. Then, RDGCN utilizes a graph attention mechanism to encourage interactions between the dual relation graph and the original graph.


\begin{figure*}[ht]
	\centering
	\subfigure[original benchmark datasets]{
		\begin{minipage}{8cm}
			\centering
			\includegraphics[width=8cm]{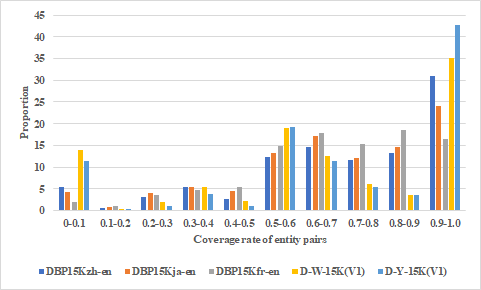}
		\end{minipage}%
	}%
	\subfigure[new benchmark datasets]{
		\begin{minipage}{8cm}
			\centering
			\includegraphics[width=8cm]{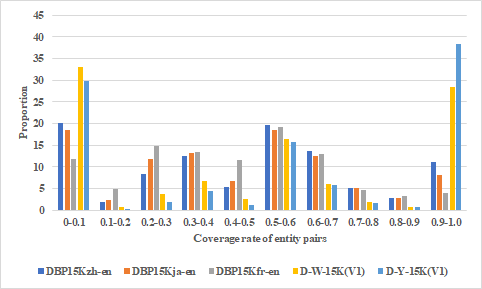}
		\end{minipage}%
	}%
	\caption{Percentage of coverage rate of entity pairs in each stage of the original benchmark datasets and new benchmark datasets. The x-axis represents the coverage rate of entity pairs, while the y-axis represents the proportion of all benchmark datasets.}
	\label{distance}
\end{figure*}

\subsection{Attribute Similarity Module}
We assume that the same entities have similar attribute value information, and the similarity of entity pairs is obtained by calculating the similarity between the attribute values they possess. 
We make the following assumption: if an attribute value is observed to occur only once in a KG, then it can be regarded as a crucial attribute value for distinguishing entities. Under this assumption, if two entities from different KGs share such a same attribute value, it is determined that these two entities should be aligned.

Formally, given a KG $G=(E, R, A, V, T_r, T_a)$ and $v_i\in V$, if attribute value $v_i$ only occurs once in $G$, such that $(e, a, v_i)\in T_a$, then we can consider $v_i$ is a crucial distinguishing attribute value. That is, if $(e_1^1, a_j^1, v_i)  \in T_a^1,(e_1^2,a_j^2, v_i)\in T_a^2 $, we can get $e_ 1^1\equiv e_1^2$.
Similarly, if the number of occurrences of attribute value $v$ is $k$, and the frequency $v_{fre} = \frac {1} {k}$, then the probability of alignment of two entities with attribute value $v$ is $\frac {1} {k} $. 
For a given entity pair $(e_1^1, e_1^2)$, we need to obtain the overlapping attribute value sets $(v_i^1, v_j^2)$ of the two entities. 
The similarity of entity pair $S(e_1^1, e_1^2) $ is calculated by $S(e_1^1, e_1^2) = \sum(v_{frei}^1, v_{frej}^2)$.

\subsection{Interaction module}
We propose two interaction approaches in the interaction module: result correction based on attribute information discrimination (RC) and matrix combination based on parameter search (PS) in order to fully combine the information obtained from encoder and attribute similarity module.

\subsubsection{Result Correction Based on Attribute Information Discrimination}

The encoder gets the entity embedding by training the labeled data, and then gets the preliminary entity alignment result by embedding distance. 
Attribute similarity module does not require any labeled data training, and attribute similarity of entities is obtained by comparing the overlapping attribute value sets. 
Because of the independence of the two modules, the two alignment results are also independent of each other. In order to effectively combine the two results, we retain the result with high confidence in the attribute similarity result, replace the result obtained by the encoder at the same position, and obtain the final result of the entity alignment framework. 


\subsubsection{Matrix Combination Based on Parameter Search} 
Inspired by a inference rule in the formal logic NAL \cite{schneider2011nexus} we propose another interaction method, which is to combine the entity embedding similarity matrix $S^{EA}$ obtained by encoder and attribute similarity matrix $S^{AT}$, and form the combined similarity matrix that used to evaluate alignment performance. 
For each pair of test entity $e_i^1 \in E^1$ and $e_j^2 \in E^2$, $S^{EA}_{ij}$ and $S^{AT}_{ij}$ can be seen as similarity belief statements, as shown in equation (3). In the truth-value, $f^{EA}$ and $f^{AT}$ represent frequency in $S^{EA}$ and $S^{AT}$ respectively, $c^{EA}$ and $c^{AT}$ represent confidence in $S^{EA}$ and $S^{AT}$ respectively. $\leftrightarrow$ represents similarity.

\begin{equation}
\begin{split}
    e_i^1\leftrightarrow e_j^2{\ }\langle f^{EA},c^{EA}\rangle\\
    e_i^1\leftrightarrow e_j^2{\ }\langle f^{AT},c^{AT}\rangle
\end{split}
\end{equation}
Then a revision rule inference can be performed to combine the two beliefs due to approximately disjoint evidence bases. 
The truth-value function is shown in equation (4), where $f_1$/$f_2$ and $c_1$/$c_2$ represents the frequency and confidence of the first/second premise belief, $f$ and $c$ represents those in the conclusion belief.
\begin{equation}
\begin{split}
    &f=\frac{f_1c_1(1-c_2)+f_2 c_2(1-c_1)}{c_1(1-c_2)+c_2(1-c_1)}\\
    &c=\frac{c_1(1-c_2)+c_2(1-c_1)}{c_1(1-c_2)+c_2(1-c_1)+(1-c_1)(1-c_2)}
\end{split}
\end{equation}

The expectation of the conclusion belief $e_i^1\leftrightarrow e_j^2{\ }\langle f,c\rangle$ is computed by $c \times (f-1/2) + 1/2$. Finally, the conclusion expectations constitute the elements of the combined similarity matrix.
The element of $S^{EA}_{ij}$ and $S^{AT}_{ij}$ corresponds to $f_{ij}^{EA}$ and $f_{ij}^{AT}$, respectively. $c^{EA}$ as a parameter represents the overall reliability (or we can call it as the expected amount of evidence to support the belief) of the entity embedding similarity matrix, and analogously $c^{AT}$ as another parameter. A parameter grid search for $c^{EA}$ and $c^{AT}$ is performed on each experiment scenario.

\begin{table}
\centering
\caption{Details of the new benchmark datasets. Original Rel. Triples represent the triplets in the original benchmark datasets.}\label{tab1}
\resizebox{\linewidth}{!}{
\begin{tabular}{cc|cccccc}
\hline
\multicolumn{2}{c|}{Datasets}  & Entities & Rel.&Attr. & Original Rel. Triples & Rel. Triples&Attr. Triples\\
\hline
DBP15K$_{hs}$ZH-EN   & ZH & 19388 & 1701&7780 & 70414 & 56530 &379684\\
        & EN & 19572 & 1323 &6933& 95142 & 85240&567755\\
      
DBP15K$_{hs}$JA-EN   & JA & 19814 & 1299 &5681& 77241 & 62737&354619\\
        & EN & 19780 & 1153 &5850& 93484 & 83526&497230\\
      
DBP15K$_{hs}$FR-EN   & FR & 19661 & 903 &4431& 105998 & 87883&528665\\
        & EN & 19993 & 1208 &6161& 115722 & 102274&576543\\
\hline        
D-W-15K(V1)$_{hs}$   & DB & 15000 & 248 &342& 38265 & 29251&68258\\
              & WD & 15000 & 169 &649& 42746 & 38022&138346\\
             
D-Y-15K(V1)$_{hs}$   & DB & 15000 & 165 &257& 30291 & 22827&71716\\
              & YG & 15000 & 28 &35& 26638 & 22368&132114\\
\hline
\end{tabular}
}
\end{table}

\begin{table*}
    \centering
    \caption{Main experimental results on DBP15K$_{hs}$, D-W-15K(V1)$_{hs}$ and D-Y-15K(V1)$_{hs}$. The bold numbers represent the best result, and the underlined  numbers represent the second place result.}
    \label{tab:booktabs}
    \resizebox{\linewidth}{!}{
    \begin{tabular}{cccccccccccccccc}
        \hline
         Models  &\multicolumn{3}{c}{ZH-EN}&\multicolumn{3}{c}{JA-EN}&\multicolumn{3}{c}{FR-EN}&\multicolumn{3}{c}{D-W-15K(V1)$_{hs}$}&\multicolumn{3}{c}{D-Y-15K(V1)$_{hs}$}\\
                &H@1&H@10&MRR&H@1&H@10&MRR&H@1&H@10&MRR&H@1&H@10&MRR&H@1&H@10&MRR\\
        \hline
            JAPE&23.2 &56.5 &-&26.5 &60.6 &-&22.8 &57.4 &-&16.1&38.5&0.237&24.4&48.2&0.327\\
		BootEA&40.7 &74.1 &0.521 &39.5 &74.4 &0.511 &41.3 &77.9 &0.537&26.1&52.8&0.349&31.1&55.0&0.392 \\ 
            AttrE&32.2 &65.3 &0.433 &28.1 &58.3 &0.385 &30.4 &65.1 &0.421&17.5&37.4&0.243&14.8&28.9&0.198 \\
            MultiKE&4.1 &7.3 &0.053 &8.2 &22.3 &0.131 &59.4 &73.6 &0.643&37.3&48.2&0.410&86.1&92.0&0.882 \\
		AliNet&47.6 &77.4 &0.579 &48.2 &80.6 &0.572 &47.4 &77.3 &0.576&26.5&47.4&0.341&31.7&51.0&0.389\\           
        \hline
        GCN-Align&23.9 &58.1 &-&24.8 &60.5 &-&24.6 &62.7 &-&18.2&42.3&0.261&21.7&40.6&0.284\\
        MRAEA&34.4 &70.4 &0.465 &37.2 &74.2 &0.498 &42.3 &79.5 &0.550 &32.6&62.8&0.428&38.1&63.2&0.471\\
        RDGCN&65.6 &81.7 &-&71.3 &86.7 &-&85.9 &94.6 &-&41.1&59.0&0.472&87.3&94.2&0.900\\
        RAGA&64.4 &84.8 &0.717&68.9 &88.8 &0.761&84.2 &95.8 &0.885&17.1&35.5&0.234&27.0&45.8&0.340\\
        \hline
        BERT-INT&\underline{80.7} &83.4 &0.818 &80.1 &87.0 &\underline{0.828} &\textbf{96.6} &\underline{97.7} &\underline{0.971}&0.6&2.1&0.001&\textbf{100}&\textbf{100}&\textbf{1} \\
        SDEA&\textbf{82.1} &\textbf{93.3} &\textbf{0.862} &77.4 & \underline{88.7} &0.814 &\underline{95.7} &\textbf{98.9} &\textbf{0.986}&\textbf{61.1}&\textbf{69.3}&\textbf{0.736}&\underline{99.4}&\underline{99.9}&\underline{0.999} \\
        \hline
        Attr-Int(RC)&73.6 &- &- &75.0 &- &- &78.1 &- &-&\underline{58.9}&-&-&97.8&-&-\\
        Attr-Int(PS)&79.7 &\underline{90.5} &\underline{0.836} &\textbf{81.8} &\textbf{91.6} &\textbf{0.853} &91.0 &96.2 &0.929&56.0&\underline{67.5}&\underline{0.600}&98.4&99.7&0.989\\
        \hline
    \end{tabular}
    }
    
\end{table*}

\section{Experiment}
\subsection{Experimental Settings}
\subsubsection{Datasets}

Table 1 outlines the statistics of new benchmark datasets obtained by lowering the coverage rate of the original benchmarks, DBP15K \cite{sun2017cross} and two monolingual datasets with sparse version in OpenEA \cite{sun2020benchmarking}. As shown in Figure 2 (a), we count the percentage of the coverage rate of the entity pairs in each stage of the original benchmark datasets DBP15K, D-W-15K(V1) and D-Y-15K(V1). As we can see, the vast majority of entity pairs have high rates of coverage, which makes it challenging to confirm KG heterogeneity. So we process the original benchmark datasets, and obtain new benchmark datasets DBP15K$_{hs}$, D-W-15K(V1)$_{hs}$, D-Y-15K(V1)$_{hs}$.
As shown in Figure 2 (b), the entity pairs with high coverage rate have diminished. The coverage of most entity pairs is concentrated below 50\%, which is particularly evident on DBP15K$_{hs}$. Due to the fact that D-W-15K(V1)$_{hs}$ and D-Y-15K(V1)$_{hs}$ are sparse datasets, entities do not have many neighbors. During the processing, we retain some identical neighbor entities. Despite this, we reduce the proportion entity pairs with a coverage rate over 90\%  from 35.2\%, 42.8\% to 28.5\%, 38.3\% in D-W-15K(V1)$_{hs}$ and D-Y-15K(V1)$_{hs}$ respectively, and increase the proportion entity pairs with a coverage rate below 10\% from 13.9\% , 11.3\%  to 33\%, 29.9\% in D-W-15K(V1)$_{hs}$ and D-Y-15K(V1)$_{hs}$.


\subsubsection{Implementation Details}
For each dataset, we split entity pairs into training, validation, and test set with ratio of 2:1:7. 
Following \cite{sun2017cross}, we use Google Translate to translate all values to English for cross-lingual datasets.

\subsection{Compared Methods}
We conduct extensive experiments to evaluate previous representative EA methods on the new benchmark datasets. 
According to the differences in the embedding modules, methods are divided into three categories: Translation-based methods, GNN-based methods and BERT-based methods.  
We selected 11 state-of-the-art EA methods, which cover different embedding modules and contain representative methods for using attribute information. \textbf{Translation-based methods:} JAPE \cite{sun2017cross}, BootEA \cite{sun2018bootstrapping}, 
AliNet \cite{sun2020knowledge}, 
AttrE \cite{trisedya2019entity},
MultiKE \cite{zhang2019multi}. \textbf{GNN-based methods:} GCN-Align \cite{wang2018cross}, MRAEA\cite{mao2020mraea}, RDGCN \cite{wu2019relation}, RAGA \cite{zhu2021raga}. \textbf{BERT-based methods:} BERT-INT \cite{tang2020bert}, 
SDEA \cite{zhong2022semantics}.



Like SDEA, we replace entity descriptions in BERT-INT with entity names since entity descriptions are not available in all benchmark datasets. Further more, RAGA performs a stable matching algorithm for 1-1 alignment after embedding. For a fair comparison, we evaluate RAGA without stable matching algorithm.

\subsection{Main Experiments}
Table 2 compares the overall performance of our method and the baseline methods. BERT-based methods gain the best results which owe to treat entity alignment as the downstream objective to fine-tune a pre-trained BERT model.
For BERT-INT, it has a strong dependency on entity name. Since FR-EN (in DBP15K$_{hs}$) and D-Y-15K(V1)$_{hs}$ include well-aligned entity names (which are extracted literally similar), it works well on these datasets as expected. BERT-INT performs poorly on D-W-15K(V1)$_{hs}$, this is because the two KGs on the datasets do not contain literally matched entity names. SDEA achieves the best results on ZH-EN (in DBP15K$_{hs}$) and D-W-15K(V1)$_{hs}$, which is mainly attributed to the design of identifying various contributions of neighbors and handling alignment of long-tail entities. As a comparison, although our method dose not use BERT to obtain information, we still achieve comparable performance with BERT-INT and SDEA. Moreover, Attr-Int(PC) outperforms BERT-INT (80.1) by 1.7 of hits@1 on JA-EN (in DBP15K$_{hs}$).

\section{Conclusions}
In this paper, we construct new benchmark datasets to closely mimic real-world EA scenarios. Then, we propose a simple and effective entity alignment framework (Attr-Int), in which innovative attribute information interaction methods can be seamlessly integrated with any embedding encoder for entity alignment, improving the performance of original entity alignment techniques. Experimental results show that our framework outperforms the state-of-the-art approaches.

\bibliographystyle{IEEEbib}
\bibliography{strings,refs}

\end{document}